\documentclass[10pt, a4paper]{article}

\usepackage[final]{lrec-coling2024} 
\usepackage{graphicx}  
\usepackage{booktabs}
\usepackage{amsmath} 
\usepackage{afterpage}
\usepackage{hyperref}

\title{Reap the Wild Wind: Detecting Media Storms in Large-Scale News Corpora}

\name{Dror K. Markus\textsuperscript{1}, Effi Levi\textsuperscript{2}, Tamir Sheafer\textsuperscript{1,3}, Shaul R. Shenhav\textsuperscript{1}} 

\address{\textsuperscript{1} Department of Political Science, The Hebrew University of Jerusalem \\ \textsuperscript{2} Institute of Computer Science, The Hebrew University of Jerusalem \\ \textsuperscript{3} Department of Communication and Journalism, The Hebrew University of Jerusalem \\ {\tt \{dror.markus|tamir.sheafer|shaul.shenhav\}@mail.huji.ac.il} \\ {\tt efle@cs.huji.ac.il}}

\abstract{Media Storms, dramatic outbursts of attention to a story, are central components of media dynamics and the attention landscape. Despite their significance, there has been little systematic and empirical research on this concept due to issues of measurement and operationalization. We introduce an iterative human-in-the-loop method to identify media storms in a large-scale corpus of news articles. The text is first transformed into signals of dispersion based on several textual characteristics. In each iteration, we apply unsupervised anomaly detection to these signals; each anomaly is then validated by an expert to confirm the presence of a storm, and those results are then used to tune the anomaly detection in the next iteration. We demonstrate the applicability of this method in two scenarios: first, supplementing an initial list of media storms within a specific time frame; and second, detecting media storms in new time periods. We make available a media storm dataset compiled using both scenarios. Both the method and dataset offer the basis for comprehensive empirical research into the concept of media storms, including characterizing them and predicting their outbursts and durations, in mainstream media or social media platforms.
 \\ \newline  }

\begin{document}

\maketitleabstract
\section{Introduction}
\label{sec:introduction}
We can recognize a \textit{media storm} when we see one - a dramatic increase in media attention to a specific issue or story for a short period of time \cite{Boydstun2014}. Such outbursts include acts of terrorism, public scandals, or discussions of provocative political decisions. They usually begin with a specific trigger event \cite[e.g.,][]{Vasterman2005, Wien2009}, and then surge to disproportionate levels of coverage - \textit{hype} \cite[e.g.,][]{AtteveldtRuigrokWelbersJacobi+2018+61+82}. Storms are central components of media dynamics, intensifying nearly all media-related effects \cite{Boydstun2014, walgraveetal2017}. In addition, being pivotal moments in the public agenda, storms can be critical junctures for political actors  \cite{Gruszcynski2020, wolfsfeld_sheafer_2006}. 

However, we still lack a systematic and comprehensive understanding of such outbursts of media attention. One reason is that it is not clear how to operationalize this concept into a concrete measurable object \citep[][518-519]{Boydstun2014}.  Many questions abound: How long must a storm last? How explosive must it be? Do all storms reach the same volume of attention? Essentially, previous researchers are left devising “arbitrary” thresholds and guidelines for their individual studies \citep[][519]{Boydstun2014}. In addition to this amorphousness, an additional challenge is that media storms are relatively sporadic phenomena.  \citet{Boydstun2014} approximate that they consist about 11\% of all media coverage, a finding that was later corroborated by~\citet{Nicholls+Bright2019}. 
The difficulty in delineating media storms and their relative rarity in target data impedes the implementation of a supervised approach to automatic storm detection. It is extremely difficult to find and annotate a gold-labeled data-set to train a model, or to even begin reading the raw articles to identify media storms directly. This makes it necessary to develop a different strategy to solve this challenging task.

Traditionally, Communication researchers employed manual content analysis to label and measure issue attention over short periods \citep[e.g.,][]{Boydstun2014, wolfsfeld_sheafer_2006}. Recent computational work has utilized topic modeling \cite{AtteveldtRuigrokWelbersJacobi+2018+61+82, nakshatri-etal-2023-using} and keyword analysis \cite{lukito-etal-2019-using} for the task. However, the drawback of these approaches is their sensitivity to research design - the keyword choice or delineation of topics. For example, a researcher might choose a topic model with topics that are too broad - hampering the ability to recognize deviances of specific outburst. Conversely, an overly complex or multidimensional topic model might cause a media storm to be dispersed across several topics. This dispersion could dilute attention peaks, making significant media events less discernible. Meanwhile, focusing on keywords may obfuscate the actual story behind the tokens.

A second approach adopted in recent computational communication research has been to focus on \textit{news story chains}. Such methods utilize clustering to identify news events - articles describing the same event or story. In some cases, the documents are converted into a network, and community detection is used \cite{Nicholls+Bright2019, Trilling+vanHoof}. In others, the clustering is preformed on document embeddings \cite{litterer-etal-2023-rains}. These techniques ‘uncover’ the stories occurring in the corpus - groups of documents discussing the same, specific event.  However, while such a method is successful at identifying stories, it is limited in the ability to detect storms as it excludes the temporal dimension. We are not simply looking for large stories, but also need to identify 'hype' - dramatic and anomalous levels of coverage of a story \cite{AtteveldtRuigrokWelbersJacobi+2018+61+82, Vasterman2005}. It is impossible to determine hype when only taking into account the structure of a single story at a single time-step without noting long-term trends and cycles as baselines.

In this work, we seek to go back to the intuitive definition of a storm - a dramatic increase in attention to an issue (above the norm) for a limited time period. Essentially, all previous research agrees on the conception of storms being anomalies of news coverage. Thus, we set out to automatically identify instances using anomaly detection. Specifically, we create several signals representing the daily dispersion of texts across the time frame. These signals are the basis for a two-step procedure. First, we utilize an unsupervised anomaly detection model to detect \textit{media storm candidates} - anomalous periods of news convergence. Second, we integrate a domain expert to determine which of the candidates should be labeled as media storms. We utilize multiple iterations of this human-in-the-loop procedure until convergence - uncovering the media storms for a period.

This approach offers several advantages. First, methodologically speaking, it integrates the temporal features of topic- or keyword-based outlier detection described above, without relying on or being limited by idiosyncrasies of researcher design. Additionally, the utilization of unsupervised anomaly detection allows us to overcome the huge quantities of data, presenting experts with a small set of candidates to focus on in determining the existence of media storms. Furthermore, our approach attempts to bypass the inherent amorphousness, offering a solution that is not built upon pre-defined statistical thresholds or ‘arbitrary’ definitions. The use of unsupervised anomaly detection allows media dynamics to reveal themselves in the data. Our expert input comes into play in validating these patterns, confirming they correspond to the theoretical concept. Thus, we are able to uncover additional, more diverse media storms than in previous studies. 

We utilize a large-scale corpus of news articles spanning 20 years of media coverage (1996-2016) to demonstrate our method. We employ two distinct experimental setups, addressing a broad spectrum of potential research applications. The first setup utilizes a seed list of media storms to uncover additional occurrences within the same time frame. The second setup utilizes an analyzed and labeled time frame to detect media storms in a new, unlabelled target period. We conclude with a preliminary analysis of our findings from both setups, underscoring the efficacy of our method and its potential for media storm research.

Finally, we contribute these findings as a media storms dataset for the years 1996-2016. We believe that this dataset opens up a wide array of exciting research avenues. While the concept of media storms holds great significance to social actors, politicians and social scientists from various fields, empirical exploration has been limited. As the classification of storms within large-scale, news coverage data improves, we can enhance our understanding of how these news hypes unfold from a single story or event to a cascade of public interest. In an era marked by heightened concern over the media's impact on the information landscape -- highlighted by issues like polarization, the spread of misinformation, and the prominence of social media -- such insights into these significant elements should offer important contributions.

\section{Data}

\subsection{News Articles}

To track the media coverage, we assembled a corpus of 1,187,607 news articles taken from three major news outlets -- the \textit{New York Times}, the \textit{Los Angeles Times} and the \textit{Washington Post} -- between 1996 and 2016. All full-length texts for this time period purchased and downloaded via a license agreement with LexisNexis.~\footnote{\url{https://www.lexisnexis.com}} These were filtered to include only articles from the News and Editorial sections. Corpus statistics are detailed in Table~\ref{tab:corpus_stats}. 

\begin{table*}[!htbp]
\centering
\begin{tabular}{lccc}
\hline
\textbf{Media Outlet} & \textbf{Articles} & \textbf{Tokens} & \textbf{Tokens/Article (Avg.)} \\
\hline
\textit{New York Times} & 520,648 & 373,980,075 & 718.30 \\
\textit{Washington Post} & 360,788 & 293,024,961 & 812.18 \\
\textit{Los Angeles Times} & 306,171 &  240,119,545 & 784.27 \\
\textbf{Total} & \textbf{1,187,607} & \textbf{907,124,581} &  \textbf{763.83} \\
\hline
\end{tabular}
\caption{Corpus Statistics}
\label{tab:corpus_stats}
\end{table*}

\subsection{Seed list of Media Storms}
\label{sec:seed list}

To initialize our method, we build upon a seed list of media storms to begin calibrating the hyperparameters of the unsupervised anomaly detection.
We begin with a list of storms from \citet{Boydstun2014} that has been widely used in media storm research. The researchers labeled the New York Times front page for a 10-year period to manually identify media storms. However, their effort contained several self-acknowledged constraints: they focused solely on domestic issues, measured only one national newspaper, and chose arbitrary statistical thresholds for operationalization. We wish to capture the essence of a media storm through a small set of mega-stories of national and global significance (expected to be present in the three outlets included in our corpus). 

Consequently, we started with the items on their list as media storm \textit{candidates}, which we could use for our first experimental setup of the method (within the 10-year period overlapping with our corpus collection: 1996-2006). However, we adjusted their list to better suit our use-case. First, since they analyzed only the New York Times, we included only national-level stories. For example, storms regarding local sports teams or municipal politics were removed. Second, we extended the list to include significant international stories, such as wars and foreign disasters, which also meet our 
conception 'media storms'. The end result is a modified list of 48 media storms between the year 1996 and 2006. We used this list to initialize the first calibration iteration of our unsupervised analysis of the full corpus in the first experimental setup (described in Section~\ref{sec:method}). We note that these are seed storm candidates used to begin the exploration of our data; we are aware that some of these events might not register as media storms after running our automated method, and that they do not represent all media storms occurring in the time period.

\section{Method}
\label{sec:method}
In this section, we present our method to detect media storms in a large corpus. First, we describe the representation of our texts into dispersion signals. Second, we detail the unsupervised anomaly detection model employed to analyze the signals. Finally, we outline the integration of the dispersion signals, anomaly detection and human-in-the-loop validation in a media storm detection method.

\subsection{Representation}
\label{sec:representation}

Our basic assumption is that during media storms, the news coverage converges surrounding a single story or event, decreasing its variance. Thus, we utilize the following method to refine the raw text into a one-dimensional signal representing the daily media dispersion. For each day in the duration of our research period, the corresponding news articles are converted into a multi-dimensional embedding. We calculate a covariance matrix based on this embedding, to capture the variance between all the day's articles over all of the embeddings' dimensions. However, since we are interested in capturing the dynamic of the dispersion over time, we calculate the commonly-used \textit{trace} value (normalized by the number of articles published that day). This provides us with a single value for the daily dispersion of the news articles. These are then aggregated to compile one-dimensional dispersion signals for the full duration of the research corpus.

In identifying media storms, we seek to include multiple representations of the texts, capturing diverse discursive attributes. We do this due to the complexity of media storms.
In some cases, they might correspond to a single event; in others, they might evolve to encompass multiple stories and news 'angles'. In some cases, such as in crises or scandals, we might expect to find specific textual styles expressing drama or surprise. However, in cases such as groundbreaking court cases or anticipated political events, the storm is signaled by the sheer volume of coverage rather than any specific reporting approach. With this complexity in mind, we incorporated four types of document embeddings to create four separate dispersion signals. This offers a level of robustness -- we seek coverage anomalies not based on any single type of textual dispersion.

\subsubsection{Actors \& Settings}
Actors are integral components of news stories. Previous research on 
automated identification of news events does so by focusing on entities, assuming
that texts referring to the same people, places and times in the same period, refer to the same news event \cite{Nicholls+Bright2019, Trilling+vanHoof}. Therefore, we include these same features in our own approach in order to identify convergence in coverage around specific events. We used the \textit{spaCy} open-source, natural language processing (NLP) named-entity recognition (NER) package \cite{honnibal2017} to extract the actors and settings of each article. For each document, we generated an embedding based on the frequency of each entity within an entity vocabulary computed over the full corpus.

\subsubsection{Topics}
In many cases, news coverage focuses more on a general issue than a specific story. For instance, strings of unrelated violent incidents could trigger a general spike in attention to crime without any of the individual events being newsworthy on their own. 
Thus, we sought to include storms being expressed in categories as opposed to only distinct stories, aligning with previous studies identifying storms as dramatic increases in coverage to an issue \cite{Boydstun2014, AtteveldtRuigrokWelbersJacobi+2018+61+82}. To generate embeddings for this feature, we utilized an unsupervised topic model -- \textit{top2vec} -- which leverages joint document and word semantic embedding to find topic vectors in a corpus \cite{angelov2020top2vec}. Such topics focus on the issues expressed in the news articles. We trained a model containing 100 topics, so each document was represented by a 100-dimensional vector. Each dimension's value was the cosine distance of the document from the corresponding topic's centroid. 

\subsubsection{Narrative plot elements}
Plot refers to "the ways in which the events and characters’ actions in a story are arranged" \cite{kukkonen2019plot}, and thus provide more information on the structure and "tellability" \cite{Shenhav2015} of stories at the heart of media storms. In order to include plot elements, we used \textit{NEAT} -- a multi-label classifier that was trained on a specially compiled dataset \cite{levi-etal-2022-detecting} to identify three plot-driven, narrative elements -- \textit{complication}, \textit{resolution}, and \textit{success}. Each document was represented by three dichotomous variables to include each of the three narrative elements.

\subsubsection{Large-Language Model (LLM)} Finally, we chose to include document embeddings based on pre-trained, transformer-based LLMs. Such models uncover latent features and patterns found within texts, and have proven to be a standard for diverse NLP tasks. We used the \textit{all-mpnet-base-v2} sentence-embedding model trained with a modified pre-trained BERT network that uses siamese and triplet network structures to derive semantically meaningful sentence embeddings that can be compared using cosine-similarity \cite{reimers-2019-sentence-bert}.

We note significant correlations between the four signals (Table \ref{tab:correlations-signals}). However, the correlations indicate that there is not a complete ‘overlap’. This attests to each signal's exclusive information. 

\begin{table}[!htbp]
\centering

\begin{tabular}{lrrrr}
\toprule
{} &  \textbf{LLM} &  \textbf{Entities} &  \textbf{Plot} \\
\midrule
\textbf{Topics         } &   0.89 &      0.92 &             0.69 \\
\textbf{LLM            } &        &      0.86 &             0.88 \\
\textbf{Entities       } &    &       &             0.70 \\
\bottomrule
\end{tabular}
\caption{Correlations between Signals}
\label{tab:correlations-signals}
\end{table}

\subsection{Unsupervised Anomaly Detection}
\label{sec:prophet}

With these media dispersion signals, we can begin the detection of anomalous convergence periods. To this end, we chose to utilize Facebook Prophet \cite{taylor2018forecasting}. Prophet is an open-source library that is conceived to be a reliable "off-the-shelf" time-series forecasting model that could be easily applicable in a variety of use cases. Prophet fits an additive regression model to a time series while including components for a linear or logistic growth curve, yearly and weekly seasonality cycles, and user-designated holidays:

\begin{equation}
y(t) = g(t) + s(t) + h(t) + \varepsilon_t
\end{equation}

Where:
\begin{itemize}
\setlength\itemsep{0em}
    \item \( g(t) \) represents the trend component.
    \item \( s(t) \) denotes the seasonal component.
    \item \( h(t) \) stands for the holiday effect at time \( t \).
    \item \( \varepsilon_t \) is the error term.
\end{itemize}

The model is fitted to the time series in question, flagging data points that significantly deviate from predicted values as anomalies. The deviation is determined by the \textit{interval width} hyperparameter -- the width of the uncertainty levels ascribed to the model. For example, a wider interval means only extreme values will be labeled anomalies. Two other hyperparameters - the \textit{changepoint prior scale} and the \textit{changepoint range} - are important for our application. The first sets the number of time-series changepoints to include in the model. The second specifies the proportion of the time series used to fit these changepoints. When working with decades worth of data, such values can significantly influence the model’s predictions. For example, a lower changepoint range means that the model takes into consideration only the early portions of the time series, while a low changepoint prior leads to decreased sensitivity to fluctuations. We chose to focus on these three hyper parameters, fine-tuning them throughout our procedure to calibrate the unsupervised anomaly detection.

\subsection{Media Storm Detection}
\label{sec:2stepprocedure}
We define a two-step procedure for identifying media storms in our corpus.

\label{sec:step1}
\textbf{Step 1}: Take as input an initial list of media storms and a target corpus of media coverage represented as described in \ref{sec:representation},\footnote{Smoothed by finding the 7-day rolling mean} to run the anomaly detection. Treating the initial input list as the ``ground truth'' for the current iteration, we evaluate the model's precision and recall as follows:

\begin{align}
\text{Precision} &= \frac{\text{D}}{\text{A}} \\
\text{Recall} &= \frac{\text{D}}{\text{S}}
\end{align}

Where \textbf{D} is the number of media storms from the initial list labeled as anomalies by the model, \textbf{A} is the total number of anomalies detected by the model and \textbf{S} is the number of media storms in the initial list.

We conduct a random search \cite{BergstraBengio2012} of the hyperparameter space, running multiple instances of the anomaly detection with varying hyperparameter values. We evaluate each instance by its precision and recall, seeking iterations with the highest scores in both metrics. In cases of ties, we prioritize recall.~\footnote{We assume that our initial storm list is but a portion of the real media storms in our target period. Therefore, we prioritize maximizing our identification of these real storms, before maximizing the sensitivity of the model.}
For the optimal instance, we examine the results of the anomaly detection, noting the dates of all periods of consecutive anomalies of at least two consecutive days. We filter these to include only the time frames where a majority out of the four dispersion signals were flagged as anomalies. This criterion was added due to the inherently ambiguous nature of media storms; we want to focus on genuine media storms and not merely statistical noise originating in the anomaly detection model or borderline instances that might be contentious among researchers. This final, filtered list is our output: a collection of anomalies -- media storm candidates.

\label{sec:step2}
\textbf{Step 2}: Takes as an input the list of media storm candidates. We apply expert validation to ascertain which candidate corresponds to a genuine media storm. For each anomaly cluster, the human expert reviewed newspaper articles from the associated dates, and cross-referenced the time frame with historical events from the corresponding dates. Only anomaly clusters found to correspond to a genuine occurrence were provided descriptive labels by our expert and added to the set of media storms.

\subsection{Experimental Setups} 
\label{sec:inperiod}
We utilized this two-step procedure in two distinct setups: In-Period and Out-Period implementations. 

\textbf{In-Period} setup. In this setup, we focused on a target period between 1996-2006, aiming to expand a seed list and detect all other storms in the time frame. We started by applying the two-step procedure described in \ref{sec:2stepprocedure} to the seed list of 48 storms described in \ref{sec:seed list} and the dispersion signals for the target years described in ~\ref{sec:representation}.
The output list of validated storms from the first iteration was saved, and then used to initialize a second iteration of the procedure. The output of this iteration became the seed of the subsequent iteration. We continuously add the validated media storms to a list of finalized media storms over all iterations. We continued the iterations until reaching convergence, defined by identifying new media storms amounting to less than 1\% of our current list of finalized media storms. We note that it can be necessary to curate the finalized list of media storms to consolidate duplicate storms. These were primarily due to small variations in the anomaly dates in each iteration that may still encapsulate a single media storm time frame.  

\textbf{Out-Period}. In this setup we utilize the two-step procedure in \ref{sec:2stepprocedure}, but begin the first step with input seed storm lists for one period, to uncover an output of occurrences in a second, unlabeled time period. Specifically, we compile data from an analyzed period together with additional, unlabeled data. As per Step 1, we use the already-labeled storms to run the random search and find the optimal anomaly detection instance. Then, we implement Step 2 on the media storm candidates for the new time period. In this way, we leverage information from a previous time frame to create a list of validated media storms for the unlabeled data. 

These two experimental setups correspond with two common research scenarios. The In-Period deployment demonstrates the ability of a researcher to leverage a handful of media storm instance that they may have identified qualitatively, to curate a comprehensive list encompassing a full target period. This challenge becomes especially pronounced when transitioning from qualitative, small-scale studies to more systematic, big-data-driven research. The Out-Period deployment demonstrates the ability of leveraging an analyzed time period to detect media storms in a new time frame. This offers promise both for expanding datasets and for predictive prospects.

\section{Results \& Discussion}

\begin{table}[!htbp]
\centering
\small
\begin{tabular}{l|rrrr}
\hline
\textbf{Iteration} & \textbf{1} & \textbf{2} & \textbf{3} & \textbf{4} \\
\hline
\textbf{Storm candidates} & 116 & 141 & 132 & 133 \\
\textbf{Storms validated} &  94 &  95 &  94 &  93 \\
\textbf{Not validated} &  22 &  46 &  38 &  40 \\
\textbf{New storms} &  71 &  18 &   4 &   1 \\
\hline

\end{tabular}
\caption{In-Period iterations}
\label{tab:period1validation}
\end{table}

\begin{table*}[!htbp]
\centering
\small 
\begin{tabular}{l|rrrrrrrrrrr}

\hline
\textbf{Year} & \textbf{2007} & \textbf{2008} & \textbf{2009} & \textbf{2010} & \textbf{2011} & \textbf{2012} & \textbf{2013} & \textbf{2014}  & \textbf{2015} & \textbf{2016} \\
\hline

\textbf{Storm candidates} & 10 & 10 & 15 & 15 & 16 & 15 & 19 & 20 & 15 & 20 \\
\textbf{Storms validated} & 6 & 9 & 12 & 11 & 12 & 14 & 14 & 16 & 13 & 13 \\
\textbf{Not validated} & 4 & 1 & 3 & 4 & 4 & 1 & 5 & 4 & 2 & 7 \\
\hline

\end{tabular}

\caption{Out-Period iterations}
\label{tab:period2validation}
\end{table*}

Table \ref{tab:period1validation} shows the results of the media storm detection in the In-Period experimental setup. There, we performed four rounds of our procedure until reaching convergence -- adding a single new media storm to our collection of 100 finalized storms. These results are noted in the table. For each round, we count the number of storm candidates found, the number of candidates validated as new media storms, and the number of candidates found to not correspond with storms, as described in \ref{sec:step1}. Additionally, since in this setup we run multiple rounds on the same period, we note the completely newly-discovered media storms -- instances that were not already detected in previous rounds.

Examining Table \ref{tab:period1validation}, we see that the anomaly detection procedure is relatively consistent, considering the stability of the storm candidates produced, storms validated and those candidates not validated over the four rounds. There are slight fluctuations. Specifically, we see the relative volatility between Rounds 1 and 2. However, by Round 3 the model is quite stable, with small changes in Round 4. Additionally,  we see here that the new instances decrease, indicating convergence. 

An analysis of correlations from the first round of the In-Period Experiment in Table \ref{tab:correlations} reveals that each signal contains exclusive information. Notably, the Plot signal shows the lowest correlations, perhaps due to the NEAT model being more discourse-grounded than vocabulary-based.

\begin{table}[!htbp]
\centering
\normalsize 
\begin{tabular}{lrrrr}
\toprule
{} &   \textbf{Entities} &   \textbf{LLM} &  \textbf{Plot} \\
\midrule
\textbf{Topics}        &       0.69 &  0.64 &           0.46 \\
\textbf{Entities}      &   &  0.72 &           0.47 \\
\textbf{LLM}           &    &   &           0.53 \\
\bottomrule
\end{tabular}
\caption{Correlations of Anomalies between Signals}
\label{tab:correlations}
\end{table}

In our implementation of the Out-Period experiment, we ran a single round of the two-step procedure (described in Section~\ref{sec:2stepprocedure}) for each year between 2007 and 2016 in our data, utilizing the media storms found in the previous nine years as seeds for detection in the final year.
For example, we utilized the media storms identified in the the In-Period experiment in the years 1997-2006 as our input to find the media storms of 2007. Then, to analyze 2008, we utilized the storms from the years 1998-2007, and so forth. 
Table \ref{tab:period2validation} displays the results from our Out-Period experiments. We see here the results of our media storm detection procedure for each year 
-- the number of media storm candidates identified followed by the numbers validated and not validated.
There are slight fluctuations in the results of each round. For example, in 2007 and 2008 we identified only 10 candidates, while reaching peaks of 20 candidates in 2014 and 2016. Additionally, there is a slight variance in the number of candidates verified as media storms (second row) and the number of candidates not corresponding to genuine storms. The existence of slight fluctuations seems reasonable; we would expect slight differences between periods when working with long-period temporal data.

\begin{table*}[!htbp]
\centering

\begin{tabular}{ccccc}
\toprule
\textbf{Year} & \textbf{No. of Storms} & \textbf{Avg. Duration} & \textbf{Duration Std.} \\
\midrule
1996 & 9 & 8.33 & 5.96 \\
1997 & 9 & 6.56 & 1.59 \\
1998 & 14 & 9.14 & 4.59 \\
1999 & 9 & 7.78 & 3.80 \\
2000 & 11 & 9.73 & 7.40 \\
2001 & 4 & 9.00 & 6.73 \\
2002 & 10 & 12.60 & 7.82 \\
2003 & 10 & 19.00 & 22.77 \\
2004 & 11 & 13.00 & 10.14 \\
2005 & 9 & 8.33 & 5.32 \\
2006 & 5 & 7.80 & 3.35 \\
\bottomrule
\end{tabular}
\caption{Storms Duration Data from 1996 to 2006}
\label{tab:storm_data_period1}
\end{table*}

\begin{table*}[!htbp]
\centering

\begin{tabular}{ccccc}
\toprule
\textbf{Year} & \textbf{No. of Storms} & \textbf{Avg. Duration} & \textbf{Duration Std.} \\
\midrule
{2007} & 7 & 10.57 & 4.04 \\
{2008} & 9 & 9.22 & 4.94 \\
{2009} & 12 & 8.50 & 5.28 \\
{2010} & 11 & 7.73 & 5.66 \\
{2011} & 12 & 8.33 & 4.66 \\
{2012} & 14 & 8.64 & 5.33 \\
{2013} & 14 & 8.79 & 4.25 \\
{2014} & 16 & 8.56 & 6.36 \\
{2015} & 13 & 10.54 & 5.50 \\
{2016} & 12 & 9.42 & 4.64 \\
\bottomrule
\end{tabular}
\caption{Storms Duration Data from 2007 to 2016}
\label{tab:storm_data_2007_2016}
\end{table*}

\begin{figure}[!htbp]
\begin{center}
\includegraphics[scale=0.20]{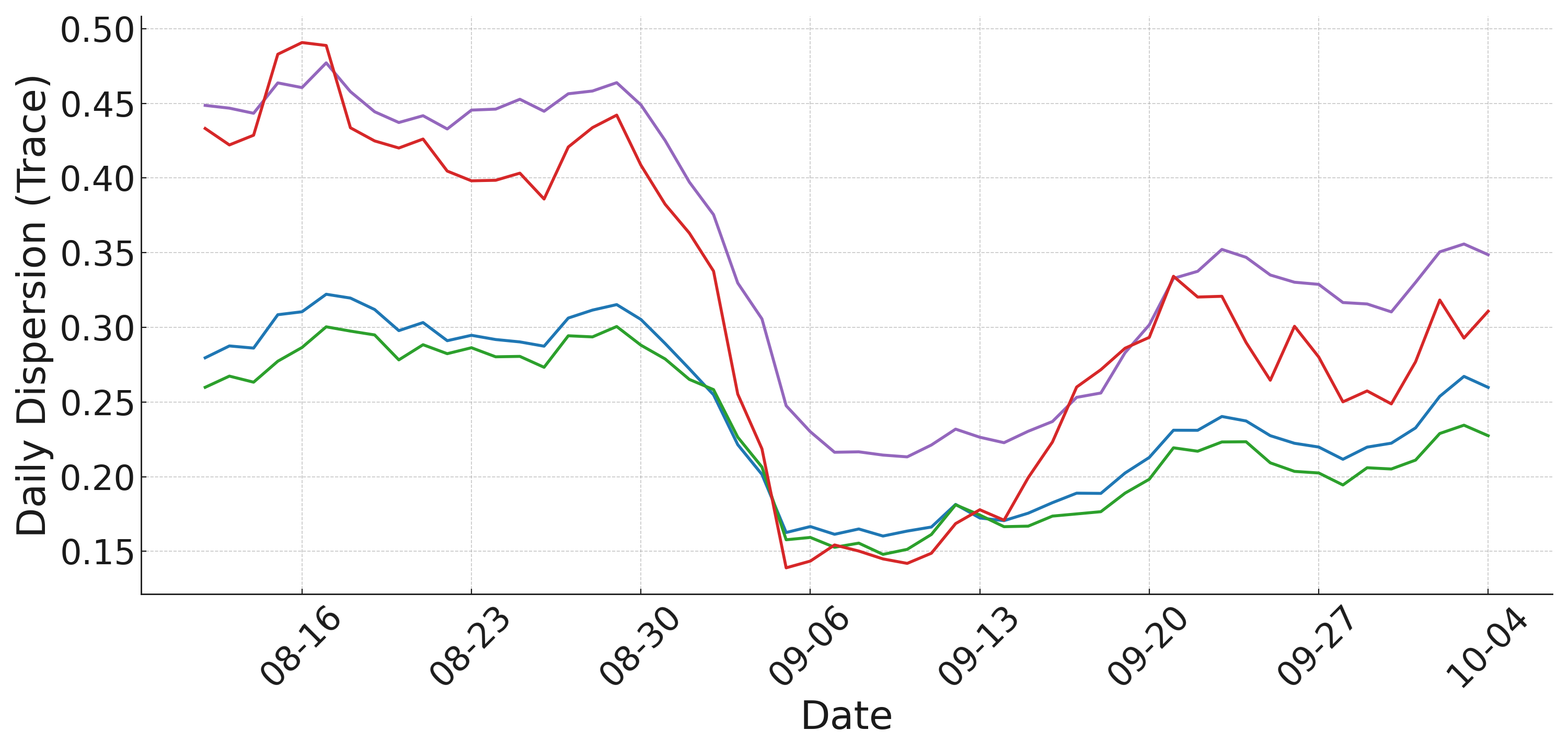} 
\caption{Hurricane Katrina -- dispersion signals: 
a visualization of the signals throughout the coverage of the hurricane. The lines correspond to LLM (purple), plot elements (red), topics (blue), and entities (green). The x-axis marks the dates and the y-axis marks the daily dispersion level (the trace).}
\label{fig:katrina}
\end{center}
\end{figure}

The end result of these experiments were 101 storms for the first period (1996-2006 - the first setup), and 120 storms for the second period (2007-2016 - the second setup) for a total of 221 media storms found in our corpus. 
These lists included many significant events, such as Hurricane Katrina (2015), the Sandy Hook school shooting and ensuing gun control debate (2012), and the Snowden NSA revelations (2013). For example, in Figure \ref{fig:katrina}, we see a graph visualizing the dispersion signals spanning the outbreak of Hurricane Katrina. The storm struck New Orleans on August 29th, leading to a humanitarian crisis. This led to a drastic convergence of coverage, expressed by a sharp decrease in the daily trace values. 

In addition to these spectacular and unanticipated events, many of the storms detected correspond to routine, planned events such as elections or sporting events. However, there were also intriguing cases such as a 2010 spike in discussion on issues of airline security and privacy. That storm does not correspond to any specific event, merely arising, we hypothesize, due to the proximity to the Thanksgiving holiday transit peak. This is an interesting example of a media storm -- public discussion of important issues -- that arises not from any specific event directly linked to the issue (We stress that this is merely a hypothesis that invites focused examination). 

In Tables \ref{tab:storm_data_period1} and \ref{tab:storm_data_2007_2016}, we see descriptive statistics for the two time periods. Overall, we see fluctuations over time in the number of storms occurring each year, and the durations (in days)
of the media storm periods. The year 2003 included 190 days of media storms, the highest in our data. This is due mostly to the U.S. war in Iraq -- including the invasion and ensuing coverage of the war's evolution (80 and 30 days, respectively).

\begin{figure}[!htbp]
\begin{center}
\includegraphics[scale=0.45]{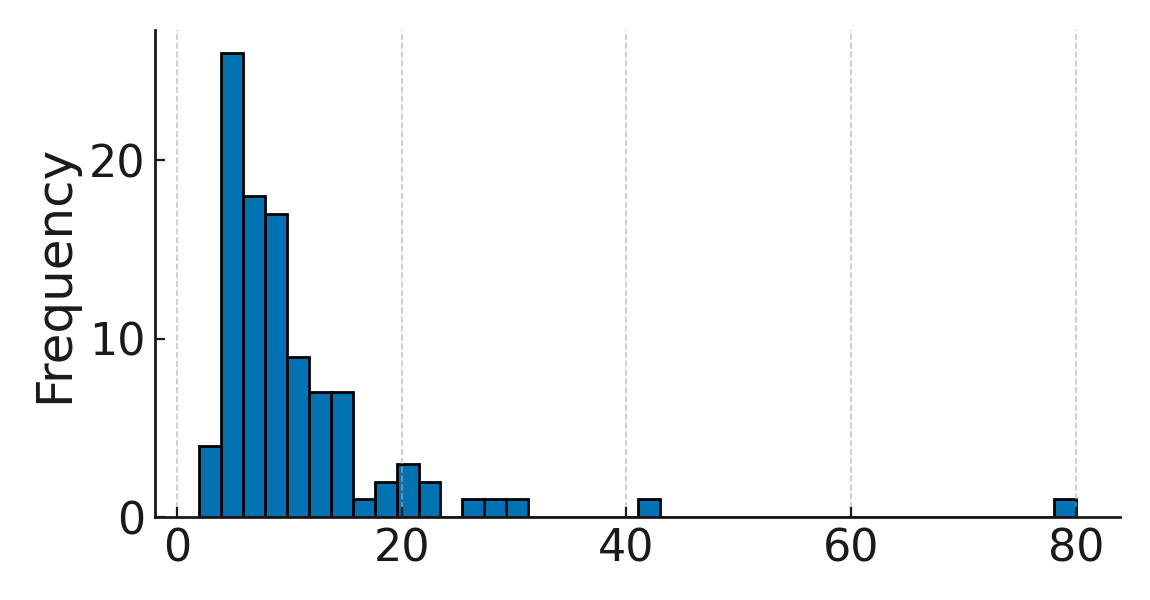} 
\vspace{-0.5cm}
\caption{Media storms durations -- "In-Period"}
\label{fig:histogram1}
\end{center}
\end{figure}

\begin{figure}[!htbp]
\begin{center}
\includegraphics[scale=0.45]{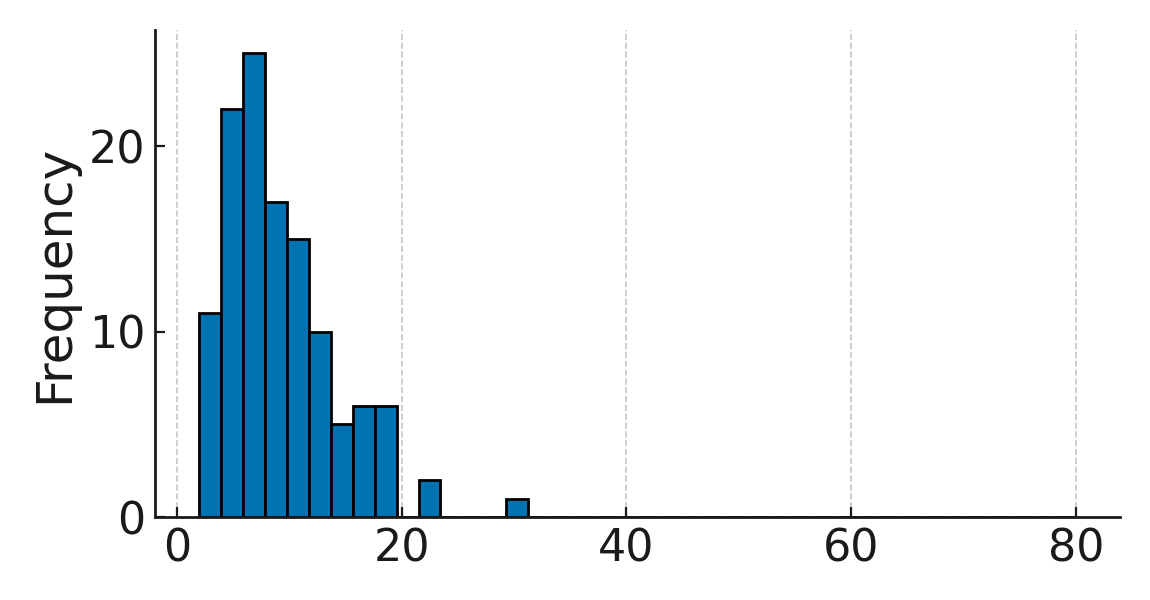} 
\vspace{-0.5cm}
\caption{Media storms durations -- "Out-Period"}
\label{fig:histogram2}
\end{center}
\end{figure}

However, what is particularly interesting about these statistics is the relative consistency of the results between the two setups. Upon examination of the results in Tables \ref{tab:storm_data_period1} and \ref{tab:storm_data_2007_2016}, we see that there are no strongly discernible differences between the media storms found in each of the setups. During the years 1996 to 2006, the annual average number of storms was $9.18$. This contrasts with the period from 2007 to 2016, which recorded an average of $12$ storms annually. This difference was statistically significant, $t(18) = -2.422$, $p = 0.026$. However, it would appear such differences might be due to real-world trends over time. 
Specifically, we see that the first years of the second period (2007 and 2008) reveal fewer storms than some of the first setup's years. Meanwhile, an examination of the storm durations (as seen in Figures \ref{fig:histogram1} and \ref{fig:histogram2}) does not reveal statistical differences (\(t(146.15) = 1.343\), \(p = 0.181\)). Such results seek to support the utility of both setups, suggesting that both are detecting the same phenomena. 

Finally, in order to understand the importance of the domain expertise, we can examine the validation statistics between the two setups: The four rounds of the the first setup found 522 media storm candidates - anomaly clusters flagged by the Prophet model. Of these, 28\% were not found to correspond to a true media storm by our human expert. The yearly rounds of storm detection in the second setup yielded a total of 155 media storm candidates, of which 22\% were deemed by the expert to not be storms. These numbers seem to justify the role of human validation.

\section{Conclusion and Future Work}

In this paper, we offer several contributions. First, we present a human-in-the-loop method to detect media storms in a large corpus of news texts. We describe a two-step iterative procedure, combining unsupervised anomaly detection and expert validation, to identify these rare events within a larger dataset. Significantly, whereas previous studies build upon 'arbitrary' statistical thresholds, we utilize the unsupervised anomaly detection to allow the media dynamics to reveal themselves in the data. Our expert input comes into play in validating these patterns, confirming they correspond to the theoretical concept. Thus, we are able to uncover additional, more nuanced media storms than in previous studies. By incorporating expert validation, we can determine the granularity of the storms we seek to identify; depending on the research application we can decide what intensity of media storms we are interested in detecting.

Second, our method offers a procedure that can be applied in various research scenarios, over diverse and large corpora, while leveraging expert knowledge for validation. Within the realm of this paper, we included three English-language newspapers for a specific time-frame. However, the method could plausibly be applied on any news corpora in any language, provided the necessary techniques could be utilized (e.g., entity-detection, sentence transformers, etc.). Additionally, researchers might be able to use this approach on non-mainstream media sources as well, including identifying periods of textual convergence in social media platforms and digital news. 

Third, through the two experimental setups, we collected a comprehensive list of media storms. This time frame we chose to focus on is of particular significance for media scholars. Between 1996 and 2016, the media landscape underwent dramatic transformations, with the rise of 24-hour news cycles, the interactivity of social media and the fragmentation of the attention landscape \cite{Chadwick_2017, edyandmerick2018}. These validated storms provide opportunities to examine intriguing theoretical questions, including how the volatility of the media landscape has evolved, changes in the events triggering storms, and perhaps developing predictive capabilities regarding storm outbursts and durations. Thus we use the results of this study to provide a dataset consisting of media storms with their start and end dates. This will be made publicly available to researchers together with the dispersion signals extracted from the corpus. 

\vspace{0.2cm}

\section{Limitations}

We note two main limitations of this project. First, the procedure described here assumes that our media storms are all mutually exclusive. We locate time frames of anomalous coverage and associate each period with a single, discrete media storm. In reality, a single time frame might contain more than one major news story, or the anomaly might actually be identified as one story declines and the other begins. Such findings correspond to issues that arose during the expert validation stage: some anomalous clusters contained a few potential storm stories. Only upon close examination of the time series' peaks and the articles that were published in correspondence with them, could we decide on a single story for the storm. Additionally, some of the periods actually did include two separate media storm stories, one following the other. In this project, we limited ourselves to choosing a single media storm per each period. In future work, however, we could integrate a clustering method to further distinguish and track stories within the media storms.

A second limitation is that our method does not include systematic steps to prevent the existence of false negatives - media storms undetected by the anomaly detection. Since we do not have a gold-standard to initiate our storm detection, there remains a possibility that our procedure may have failed to detect instances within our corpus. In general, our approach relies on high-quality seeds to initiate the search for additional media storms. We assume that these instances fully represent the phenomenon, and that, therefore, all media storms should be similar enough in characteristic to them. In this way, multiple iterations of anomaly detection should uncover all true media storms. However, we note that this is not a complete solution to the issue of false negatives. In future work, we would examine potential solutions, such as randomly sampling the non-storm time periods to examine for storms, or perhaps generating additional textual signals which might reveal more storm instances.

\section{Acknowledgements}

We thank Guy Mor-Lan for his input, and Noa Amir and Hagar Kaminer for their assistance. 

\section*{Funding}
This work was supported by a grant from the Israel Science Foundation (2501/22), as well as from the Hebrew University Center for Interdisciplinary Data Science Research (CIDR) and the Annenberg Foundation Grant for American Studies at the Hebrew University.

\nocite{*}
\section{Bibliographical References}\label{sec:reference}

\bibliographystyle{lrec-coling2024-natbib}
\bibliography{biblio}

\begin{thebibliography}{32}
\expandafter\ifx\csname natexlab\endcsname\relax\def\natexlab#1{#1}\fi

\bibitem[{Angelov(2020)}]{angelov2020top2vec}
Dimo Angelov. 2020.
\newblock \href {http://arxiv.org/abs/2008.09470} {Top2vec: Distributed representations of topics}.

\bibitem[{Barberá et~al.(2019)Barberá, Casas, Nagler, Egan, Bonneau, Jost, and Tucker}]{barberá_casas_nagler_egan_bonneau_jost_tucker_2019}
Pablo Barberá, Andreu Casas, Jonathan Nagler, Patrick~J. Egan, Richard Bonneau, John~T. Jost, and Joshua~A. Tucker. 2019.
\newblock \href {https://doi.org/10.1017/S0003055419000352} {Who leads? who follows? measuring issue attention and agenda setting by legislators and the mass public using social media data}.
\newblock \emph{American Political Science Review}, 113(4):883–901.

\bibitem[{Bergstra and Bengio(2012)}]{BergstraBengio2012}
J.~Bergstra and Y.~Bengio. 2012.
\newblock Random search for hyper-parameter optimization.
\newblock \emph{Journal of Machine Learning Research}, 13(Feb):281--305.

\bibitem[{Boydstun et~al.(2014)Boydstun, Hardy, and Walgrave}]{Boydstun2014}
Amber~E. Boydstun, Anne Hardy, and Stefaan Walgrave. 2014.
\newblock \href {https://doi.org/10.1080/10584609.2013.875967} {Two faces of media attention: Media storm versus non-storm coverage}.
\newblock \emph{Political Communication}, 31(4):509--531.

\bibitem[{Chadwick(2017)}]{Chadwick_2017}
Andrew Chadwick. 2017.
\newblock \emph{The Hybrid Media System: Politics and Power}.
\newblock Oxford University press.

\bibitem[{Ciampaglia et~al.(2015)Ciampaglia, Flammini, and Menczer}]{ciampaglia}
Giovanni~Luca Ciampaglia, Alessandro Flammini, and Filippo Menczer. 2015.
\newblock \href {https://doi.org/10.1038/srep09452} {The production of information in the attention economy}.
\newblock \emph{Scientific Reports}, 5(1):9452.

\bibitem[{Edy and Meirick(2018)}]{edyandmerick2018}
Jill~A Edy and Patrick~C Meirick. 2018.
\newblock \href {https://doi.org/10.1093/poq/nfy043} {{The Fragmenting Public Agenda: Capacity, Diversity, and Volatility in Responses to the “Most Important Problem” Question}}.
\newblock \emph{Public Opinion Quarterly}, 82(4):661--685.

\bibitem[{Galtung and Ruge(1965)}]{galtungruge}
Johan Galtung and Mari~Holmboe Ruge. 1965.
\newblock The structure of foreign news: The presentation of the congo, cuba and cyprus crises in four norwegian newspapers.
\newblock \emph{Journal of Peace Research}, 2(1):64--90.

\bibitem[{Gruszczynski(2020)}]{Gruszcynski2020}
Mike Gruszczynski. 2020.
\newblock \href {https://ijoc.org/index.php/ijoc/article/view/13267} {How media storms and topic diversity influence agenda fragmentation}.
\newblock \emph{International Journal of Communication}, 14(0).

\bibitem[{Harcup and O'Neill(2017)}]{oneill2017}
Tony Harcup and Deirdre O'Neill. 2017.
\newblock \href {https://doi.org/10.1080/1461670X.2016.1150193} {What is news?}
\newblock \emph{Journalism Studies}, 18(12):1470--1488.

\bibitem[{Honnibal and Montani(2017)}]{honnibal2017}
Matthew Honnibal and Ines Montani. 2017.
\newblock {spaCy 2}: Natural language understanding with {B}loom embeddings, convolutional neural networks and incremental parsing.
\newblock To appear.

\bibitem[{Kukkonen(2014)}]{kukkonen2019plot}
Karin Kukkonen. 2014.
\newblock \href {http://www.lhn.uni-hamburg.de/article/plot} {Plot}.
\newblock In Peter Hühn et~al., editors, \emph{The Living Handbook of Narratology}. Hamburg University, Hamburg.
\newblock Accessed: 12 Feb 2019.

\bibitem[{Levi et~al.(2022)Levi, Mor, Sheafer, and Shenhav}]{levi-etal-2022-detecting}
Effi Levi, Guy Mor, Tamir Sheafer, and Shaul Shenhav. 2022.
\newblock \href {https://doi.org/10.18653/v1/2022.findings-naacl.133} {Detecting narrative elements in informational text}.
\newblock In \emph{Findings of the Association for Computational Linguistics: NAACL 2022}, pages 1755--1765, Seattle, United States. Association for Computational Linguistics.

\bibitem[{Litterer et~al.(2023)Litterer, Jurgens, and Card}]{litterer-etal-2023-rains}
Benjamin Litterer, David Jurgens, and Dallas Card. 2023.
\newblock \href {https://doi.org/10.18653/v1/2023.findings-emnlp.420} {When it rains, it pours: Modeling media storms and the news ecosystem}.
\newblock In \emph{Findings of the Association for Computational Linguistics: EMNLP 2023}, pages 6346--6361, Singapore. Association for Computational Linguistics.

\bibitem[{Lukito et~al.(2019)Lukito, K~Sarma, Foley, and Abhishek}]{lukito-etal-2019-using}
Josephine Lukito, Prathusha K~Sarma, Jordan Foley, and Aman Abhishek. 2019.
\newblock \href {https://doi.org/10.18653/v1/W19-2107} {Using time series and natural language processing to identify viral moments in the 2016 {U}.{S}. presidential debate}.
\newblock In \emph{Proceedings of the Third Workshop on Natural Language Processing and Computational Social Science}, pages 54--64, Minneapolis, Minnesota. Association for Computational Linguistics.

\bibitem[{McCombs and Zhu(1995)}]{mccombs&zhu}
Maxwell McCombs and Jian-Hua Zhu. 1995.
\newblock \href {http://www.jstor.org/stable/2749592} {Capacity, diversity, and volatility of the public agenda: Trends from 1954 to 1994}.
\newblock \emph{The Public Opinion Quarterly}, 59(4):495--525.

\bibitem[{Nakshatri et~al.(2023)Nakshatri, Liu, Chen, Roth, Goldwasser, and Hopkins}]{nakshatri-etal-2023-using}
Nishanth Nakshatri, Siyi Liu, Sihao Chen, Dan Roth, Dan Goldwasser, and Daniel Hopkins. 2023.
\newblock \href {https://doi.org/10.18653/v1/2023.findings-emnlp.274} {Using {LLM} for improving key event discovery: Temporal-guided news stream clustering with event summaries}.
\newblock In \emph{Findings of the Association for Computational Linguistics: EMNLP 2023}, pages 4162--4173, Singapore. Association for Computational Linguistics.

\bibitem[{Nicholls and Bright(2019)}]{Nicholls+Bright2019}
Tom Nicholls and Jonathan Bright. 2019.
\newblock \href {https://doi.org/10.1080/19312458.2018.1536972} {Understanding news story chains using information retrieval and network clustering techniques}.
\newblock \emph{Communication Methods and Measures}, 13(1):43--59.

\bibitem[{Otto and Meyer(2012)}]{Otto&Meyer}
Florian Otto and Christoph~O Meyer. 2012.
\newblock \href {https://doi.org/10.1177/1750635212458621} {Missing the story? changes in foreign news reporting and their implications for conflict prevention}.
\newblock \emph{Media, War \& Conflict}, 5(3):205--221.

\bibitem[{Paimre and Harro-Loit(2018)}]{PaimreHarroLoit+2018+267+288}
Marianne Paimre and Halliki Harro-Loit. 2018.
\newblock \href {https://doi.org/doi:10.1515/9789048532100-015} {12. news waves generating attentionscapes: Opportunity or a waste of public time?}
\newblock In Peter Vasterman, editor, \emph{From Media Hype to Twitter Storm: News Explosions and Their Impact on Issues, Crises and Public Opinion}, pages 267--288. Amsterdam University Press, Amsterdam.

\bibitem[{Reimers and Gurevych(2019)}]{reimers-2019-sentence-bert}
Nils Reimers and Iryna Gurevych. 2019.
\newblock \href {https://arxiv.org/abs/1908.10084} {Sentence-bert: Sentence embeddings using siamese bert-networks}.
\newblock In \emph{Proceedings of the 2019 Conference on Empirical Methods in Natural Language Processing}. Association for Computational Linguistics.

\bibitem[{Shenhav(2015)}]{Shenhav2015}
S.R. Shenhav. 2015.
\newblock \emph{Analyzing Social Narratives}.
\newblock Routledge, New York.

\bibitem[{Swanson(2003)}]{Swanson2003}
D.L. Swanson. 2003.
\newblock Political news in the changing environment of political journalism.
\newblock In P.J. Maarek and G.~Wolfsfeld, editors, \emph{Political Communication in a New Era}, pages 11--31. Routledge Taylor and Francis Group.

\bibitem[{Taylor and Letham(2018)}]{taylor2018forecasting}
Sean~J Taylor and Benjamin Letham. 2018.
\newblock Forecasting at scale.
\newblock \emph{The American Statistician}, 72(1):37--45.

\bibitem[{Trilling and van Hoof(2020)}]{Trilling+vanHoof}
Damian Trilling and Marieke van Hoof. 2020.
\newblock \href {https://doi.org/10.1080/21670811.2020.1839352} {Between article and topic: News events as level of analysis and their computational identification}.
\newblock \emph{Digital Journalism}, 8(10):1317--1337.

\bibitem[{van Atteveldt et~al.(2018)van Atteveldt, Ruigrok, Welbers, and Jacobi}]{AtteveldtRuigrokWelbersJacobi+2018+61+82}
Wouter van Atteveldt, Nel Ruigrok, Kasper Welbers, and Carina Jacobi. 2018.
\newblock \href {https://doi.org/doi:10.1515/9789048532100-005} {2. news waves in a changing media landscape 1950-2014}.
\newblock In Peter Vasterman, editor, \emph{From Media Hype to Twitter Storm: News Explosions and Their Impact on Issues, Crises and Public Opinion}, pages 61--82. Amsterdam University Press, Amsterdam.

\bibitem[{Vasterman(2005)}]{Vasterman2005}
Peter~L.M. Vasterman. 2005.
\newblock \href {https://doi.org/10.1177/0267323105058254} {Media-hype: Self-reinforcing news waves, journalistic standards and the construction of social problems}.
\newblock \emph{European Journal of Communication}, 20(4):508--530.

\bibitem[{Waisbord and Russell(2020)}]{waisbord+russell}
Silvio Waisbord and Adrienne Russell. 2020.
\newblock \href {https://doi.org/10.1177/1077699020917116} {News flashpoints: Networked journalism and waves of coverage of social problems}.
\newblock \emph{Journalism \& Mass Communication Quarterly}, 97(2):376--392.

\bibitem[{Walgrave et~al.(2017)Walgrave, Boydstun, Vliegenthart, and Hardy}]{walgraveetal2017}
Stefaan Walgrave, Amber~E. Boydstun, Rens Vliegenthart, and Anne Hardy. 2017.
\newblock \href {https://doi.org/10.1080/10584609.2017.1289288} {The nonlinear effect of information on political attention: Media storms and u.s. congressional hearings}.
\newblock \emph{Political Communication}, 34(4):548--570.

\bibitem[{Wien and Elmelund-Præstekær(2009)}]{Wien2009}
Charlotte Wien and Christian Elmelund-Præstekær. 2009.
\newblock \href {https://doi.org/10.1177/0267323108101831} {An anatomy of media hypes: Developing a model for the dynamics and structure of intense media coverage of single issues}.
\newblock \emph{European Journal of Communication}, 24(2):183--201.

\bibitem[{Wolfsfeld(2011)}]{Wolfsfeld2011}
G.~Wolfsfeld. 2011.
\newblock \emph{Making Sense of Media and Politics: Five Principles in Political Communication}.
\newblock Routledge, New York.

\bibitem[{Wolfsfeld and Sheafer(2006)}]{wolfsfeld_sheafer_2006}
Gadi Wolfsfeld and Tamir Sheafer. 2006.
\newblock \href {https://doi.org/10.1080/10584600600808927} {Competing actors and the construction of political news: The contest over waves in israel}.
\newblock \emph{Political Communication}, 23(3):333--354.

\end{thebibliography}

\end{document}